\pdfoutput=1

\documentclass[11pt]{article}

\usepackage{acl}

\usepackage{times}
\usepackage{latexsym}

\usepackage[T1]{fontenc}

\usepackage[utf8]{inputenc}

\usepackage{microtype}

\usepackage{inconsolata}

\usepackage{amsfonts}
\usepackage{amsmath}
\usepackage{booktabs}
\usepackage{multirow}
\usepackage{graphicx}
\usepackage{enumitem}
\usepackage{textcomp}
\usepackage{booktabs}
\usepackage{subfigure}
\usepackage{array}
\usepackage{bm}
\usepackage{tabularx}
\usepackage{tabulary}
\usepackage{array}
\usepackage{color,ulem}

\title{MiLe Loss: a New Entropy-Weighed Loss for Mitigating the Bias of Learning Difficulties in Large Language Models}

\author{Zhenpeng Su\textsuperscript{\rm 1,2}\footnotemark[1], Xing Wu\textsuperscript{\rm 1,2}\footnotemark[1], Xue Bai\textsuperscript{\rm 3}\footnotemark[1], Zijia Lin\textsuperscript{\rm 3}\footnotemark[1], Hui Chen\textsuperscript{\rm 4}, Guiguang Ding\textsuperscript{\rm 4}, Wei Zhou\textsuperscript{\rm 1,2}, Songlin Hu\textsuperscript{\rm 1,2}\footnotemark[2] \\
  \textsuperscript{\rm 1}Institute of Information Engineering, Chinese Academy of Sciences, Beijing, China\\
  \textsuperscript{\rm 2}School of Cyber Security, University of Chinese Academy of Sciences, Beijing, China\\
  \textsuperscript{\rm 3}Kuaishou Technology, Beijing, China.
  \textsuperscript{\rm 4}Tsinghua University, Beijing, China\\
  \texttt{\{suzhenpeng,wuxing,zhouwei,husonglin\}@iie.ac.cn} \\
  \texttt{baixue05@kuaishou.com, linzijia07@tsinghua.org.cn} \\
  \texttt{huichen@mail.tsinghua.edu.cn, dinggg@tsinghua.edu.cn} \\ 
  }

\begin{document}
\maketitle
\renewcommand{\thefootnote}{\fnsymbol{footnote}} 
\footnotetext[1]{These authors contributed equally to this work.} 
\footnotetext[2]{Corresponding authors.} 
\renewcommand{\thefootnote}{\arabic{footnote}}
\begin{abstract}
Generative language models are usually pretrained on large text corpus via predicting the next token (i.e., sub-word/word/phrase) given the previous ones. Recent works have demonstrated the impressive performance of large generative language models on downstream tasks. However, existing generative language models generally neglect an inherent challenge in text corpus during training, i.e., the imbalance between frequent tokens and infrequent ones.
It can lead a language model to be dominated by common and easy-to-learn tokens, thereby overlooking the infrequent and difficult-to-learn ones.
To alleviate that, we propose a \textbf{MiLe Loss} function for \textbf{mi}tigating the bias of \textbf{le}arning difficulties with tokens. 
During training, it can dynamically assess the learning difficulty of a to-be-learned token, according to the information entropy of the corresponding predicted probability distribution over the vocabulary. Then it scales the training loss adaptively, trying to lead the model to focus more on the difficult-to-learn tokens. On the Pile dataset, we train generative language models at different scales of 468M, 1.2B, and 6.7B parameters. Experiments reveal that models incorporating the proposed MiLe Loss can gain consistent performance improvement on downstream benchmarks.
\end{abstract}

\section{Introduction}
Generative language models like GPT-3~\cite{DBLP:conf/nips/BrownMRSKDNSSAA20} are generally pretrained on extensive textual data, in the manner of predicting the next token given the previous ones for each training text. Recently, large generative language models have been exhibiting impressive performance on various downstream natural language tasks, like dialogue system, classification, sequence labeling, etc.~\cite{DBLP:journals/corr/abs-2302-13971, DBLP:conf/nips/BrownMRSKDNSSAA20,DBLP:journals/corr/abs-2204-02311}, and attracting much attention from both academia and industry. 

\begin{table}[!t]
\centering
\scalebox{0.9}{
\begin{tabular}{c|ccc}
\toprule
Frequency Bucket & high & medium  & low  \\
\midrule
PPL  & 4.323  & 13.541 & 15.517  \\
\bottomrule
\end{tabular}
}
\caption{The average perplexity (PPL) for tokens in different frequency buckets.}
\label{pre-expertiment}
\end{table}

However, previous works have overlooked an inherent issue in natural language corpus that might affect the pretraining of a language model, i.e., frequent tokens far outnumber infrequent ones.
Actually, Zipf's law~\cite{piantadosi2014zipf} highlights the inherent imbalance of tokens in natural language datasets, i.e., a few frequent tokens would dominate a dataset while many infrequent ones only form a minor portion.
For instance, $50\%$ of the Brown Corpus ~\cite{francis1979brown}, which comprises over a million tokens, is covered by only the top 135 most frequent tokens.

The imbalance of tokens is essentially a class imbalance problem.
We argue that infrequent tokens are difficult to learn due to their fewer occurrences, in contrast to the frequent ones that can be learned adequately~\cite{DBLP:conf/iccv/LinGGHD17}. To confirm that, we utilize the remarkable language model LLaMA~\cite{DBLP:journals/corr/abs-2302-13971} with 6.7B parameters on the Pile~\cite{DBLP:journals/corr/abs-2101-00027} validation set and perform a detailed perplexity (PPL) analysis at the token level. It's worth noting that a higher perplexity is indicative of a token's higher learning difficulty. In our analysis, all tokens are grouped into three frequency buckets: high, medium, and low, based on their counts in the whole Pile dataset\footnote{As the Pile dataset is large enough, the relative frequencies of all tokens are supposed to be almost the same as those in the training set of LLaMA, which is not publicly available.}. 
Here, we calculate the frequency of each token and sort them in descending order of frequency. Then, we categorize the top tokens that cover $80\%$ of the dataset as tokens of high frequency, those that cover the extra $15\%$ (i.e., $80\%-95\%$) of the dataset as tokens of medium frequency, and the remaining $5\%$ as tokens of low frequency.
As shown in Table \ref{pre-expertiment}, for the tokens of high frequency, LLaMA derives a much lower average perplexity ($4.394$) than those of medium ($13.891$) or low ($15.814$) frequency.
That confirms our assumption: token imbalance can lead to the bias of learning difficulties.
More explicitly, those frequent and easy-to-learn tokens (i.e., classes) might overwhelm the model and make it neglect the infrequent and difficult-to-learn ones during training~\cite{DBLP:conf/iccv/LinGGHD17}. Therefore, we emphasize that the latter kinds of tokens should be given more attention during language model pretraining. 



\begin{figure}[!t]
    \centering
    \includegraphics[width=0.5\textwidth]{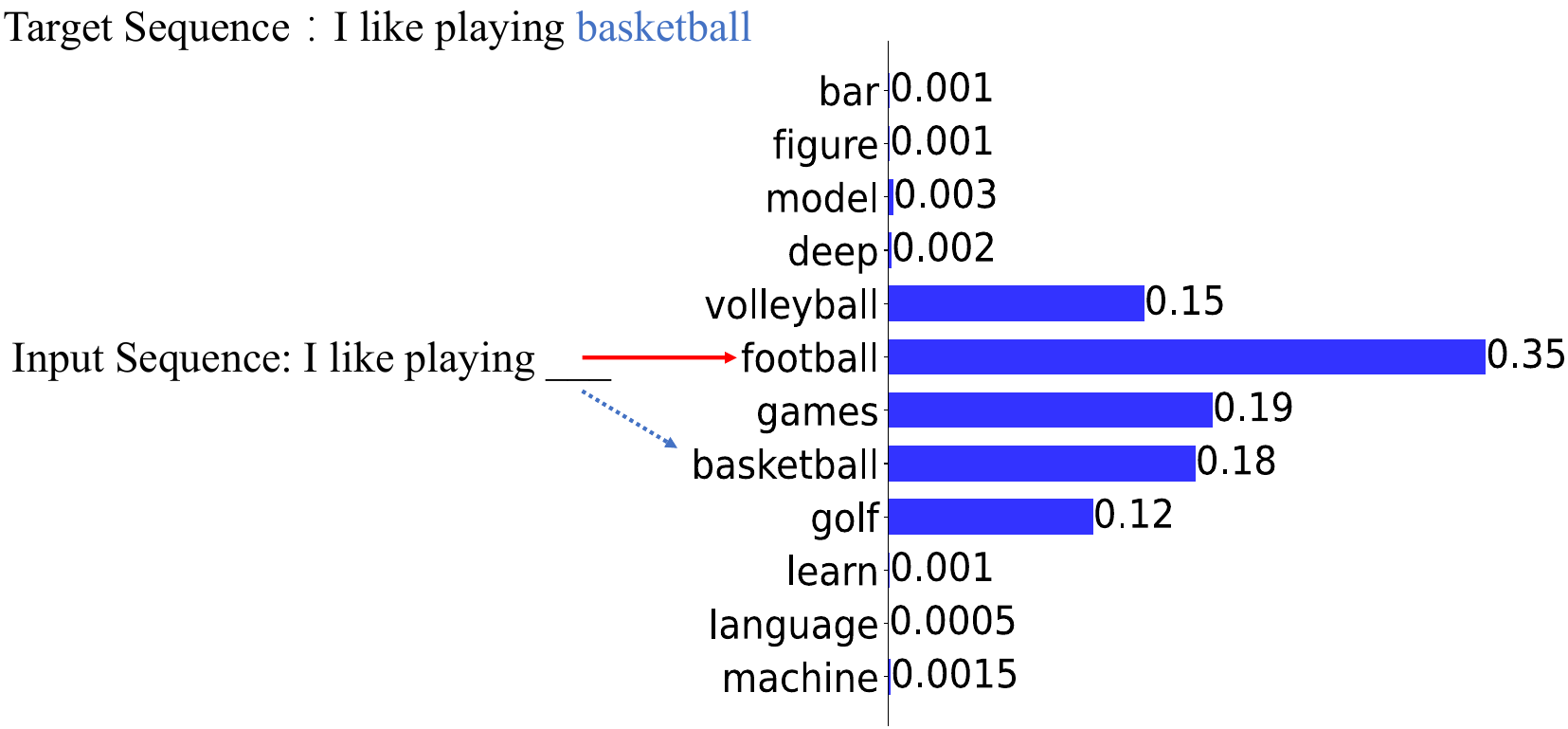}
    \caption{An example where predicting the next token is more like a multi-label classification problem.}
    \label{fig:exaple}
\end{figure}

It is a straightforward idea to use the notable Focal Loss~\cite{DBLP:conf/iccv/LinGGHD17} from the field of object detection as an alternative to the prevalent Cross-Entropy Loss for the next token prediction. 
This modification aims to intensify the language model's focus on the infrequent and difficult-to-learn tokens.
Focal Loss is a dynamically scaled version of the Cross-Entropy Loss, where the scaling factor decreases as the predicted probability w.r.t the ground-truth token increases. 
Specifically, Focal Loss decreases the weights of the easy-to-learn tokens, as their predicted probabilities are higher, and meanwhile increases the weights of the difficult-to-learn ones, as their predicted probabilities are lower. In that way, it compels the language model to pay more attention to difficult-to-learn tokens.

Nevertheless, Focal Loss~\cite{DBLP:conf/iccv/LinGGHD17} only takes into account the probability w.r.t the ground-truth token when assessing its learning difficulty, and is intuitively designed for the multi-class classification problem where an object is only associated with a single class label.  
Indeed, in language model pretraining, when predicting the next token given the previous ones, there might exist multiple valid tokens besides the ground-truth one. 
This makes predicting the next token more like a multi-label classification problem, where an object can be associated with multiple class labels~\cite{tsoumakas2007multi,DBLP:conf/ijcai/ChenDLZH18}. 
For example, as shown in Figure \ref{fig:exaple}, given the previous tokens ``I like playing '', there are multiple valid next tokens, like ``basketball'', ``football'', ``golf'', etc. 
Suppose the target training token sequence is ``I like playing basketball''. 
As the valid tokens would divide up almost the total probability (i.e., 1.0), the ground-truth token ``basketball'' would be given a smaller probability (e.g., 0.18).
Then Focal Loss would treat ``basketball'' for the position as a difficult-to-learn token.
However, as all the other valid tokens are also correct for the position in the view of language modeling, only allowing ``basketball'' to be predicted is unsuitable. Thus, the learning difficulty assessed by Focal Loss for ``basketball'' is imperfect in such a multi-label classification case. 

In this paper, we propose a new loss function termed MiLe Loss to better enable a language model to pay more attention to the difficult-to-learn tokens in such multi-label classification cases.
We observe that when a next target token is easy-to-learn, the minor valid tokens would divide up almost the total probability while others are associated with very low probabilities, resulting in a low information entropy of the predicted probability distribution over the vocabulary. On the contrary, if a next token is difficult-to-learn, the predicted probability distribution would be more uniform, resulting in a higher information entropy. 
Therefore, instead of relying on the single probability of the ground-truth token as Focal Loss, the proposed MiLe Loss uses the information entropy of the predicted probability distribution for assessing learning difficulties, which can better handle cases with multiple valid tokens.
Then, tokens exhibiting high-entropy, possibly being difficult-to-learn, will be assigned increased weights during language model pretraining. 

To validate the effectiveness of the proposed MiLe Loss, we train three different-sized models on the Pile dataset~\cite{DBLP:journals/corr/abs-2101-00027}. 
Experimental results indicate that MiLe Loss steadily outperforms Focal Loss and Cross-Entropy Loss on downstream benchmarks.

Our contributions can be summarized as follows.

\begin{itemize}[leftmargin=*]
    \item We highlight the bias of learning difficulties in generative language models, which is mainly caused by the inherent token imbalance in textual training data.
    \item We propose a new loss function termed MiLe Loss to enhance Focal Loss for mitigating the bias of learning difficulties.
    \item  We validate the effectiveness of the proposed MiLe Loss with extensive experiments. Experimental results show that it consistently outperforms Focal Loss and Cross-Entropy Loss.
\end{itemize}

\section{Related Works}
\subsection{Language Models}
Language Models are statistical models that aim to maximize the likelihood of the training sequences of tokens~\cite{DBLP:journals/corr/abs-2302-13971}.
Early language models are based on the statistics of $n$-grams~\cite{DBLP:journals/pami/BahlJM83,DBLP:journals/tsp/Katz87,DBLP:conf/icassp/KneserN95}.
Then the focus has shifted toward neural-network-based models. Recurrent Neural Networks~\cite{DBLP:conf/interspeech/MikolovKBCK10} and their variants, e.g., LSTMs~\cite{DBLP:journals/corr/Graves13}, have been successful in this regard. Those models are capable of learning complex patterns in textual data and have achieved remarkable results in various language modeling tasks.

Recently, Transformers are commonly used as the backbone network for language models. Representative works include BERT~\cite{DBLP:conf/naacl/DevlinCLT19}, RoBERTa~\cite{DBLP:journals/corr/abs-1907-11692}, GPT-2~\cite{radford2019language}, UniLM~\cite{DBLP:conf/nips/00040WWLWGZH19}, and T5~\cite{DBLP:journals/jmlr/RaffelSRLNMZLL20}, etc. Since the advent of GPT-3~\cite{DBLP:conf/nips/BrownMRSKDNSSAA20} with 175 billion parameters, which achieves outstanding performance in various downstream tasks, the research landscape has increasingly pivoted towards large generative language models. Notable works like Gopher~\cite{DBLP:journals/corr/abs-2112-11446}, Pythia~\cite{DBLP:conf/icml/BidermanSABOHKP23}, PaLM~\cite{DBLP:journals/corr/abs-2204-02311}, GaLM~\cite{DBLP:conf/icml/DuHDTLXKZYFZFBZ22}, OPT~\cite{zhang2022opt} and LLaMA~\cite{DBLP:journals/corr/abs-2302-13971}, have also been proposed. 

However, previous works do not consider the bias of learning difficulties among tokens, which is mainly caused by the inherent token imbalance in the textual training data. 
They probably overlook some difficult-to-learn but informative tokens during model training. 
To tackle that, in this paper we introduce MiLe Loss, aiming to lead generative language models to pay more attention to those tokens.

\subsection{Class Imbalance}
Class Imbalance refers to a highly skewed distribution of classes in the training data, which means that the number of instances in some classes is significantly higher than those in the other classes~\cite{yang2020rethinking}. A commonly used solution is to perform data re-sampling, where the minority classes are up-sampled~\cite{DBLP:journals/jair/ChawlaBHK02,DBLP:conf/pkdd/AndoH17,DBLP:conf/mipr/PouyanfarTMTKGD18,shen2016relay}, and the majority classes are down-sampled~\cite{DBLP:conf/icip/LeePK16, DBLP:journals/nn/BudaMM18}.
Other works~\cite{DBLP:conf/cvpr/CuiJLSB19, DBLP:journals/pami/DongGZ19, DBLP:conf/iccv/LinGGHD17} have also proposed enhanced loss functions to mitigate issues caused by class imbalance, e.g., Focal Loss. 


In language modeling, to mitigate the mentioned bias of learning difficulties caused by the inherent token imbalance, one may simply refer to the data re-sampling method.  
However, data re-sampling at the \texttt{token} level, i.e., up-sampling infrequent tokens and down-sampling frequent ones, will probably break the semantics of training texts. 
Meanwhile, re-sampling at the coarse-grained \texttt{sentence}/\texttt{paragraph}/\texttt{document}/\texttt{domain} level will equally increase/decrease the number of both kinds of tokens, and thus cannot well tackle the token imbalance.

Therefore, we consider enhancing the loss function to alleviate the bias of learning difficulties among tokens for generative language models, enabling them to pay more attention to those difficult-to-learn but informative tokens. Firstly, we attempted to use the notable Focal Loss. However, since predicting the next token in generative language models is more like a multi-label classification problem as analyzed before, Focal Loss struggles to give suitable scaling factors for cases with multiple valid next tokens. To tackle that, we introduce the MiLe Loss. 

\section{Method}

\subsection{Preliminaries}

\paragraph{\textbf{Language Model Pretraining}}  As mentioned before, a generative language model is generally trained via predicting the next token (i.e., sub-word/word/phrase), one by one, based on the previous ones for each training text, aiming to maximize the likelihood. 
Formally, given a training text $T$ consisting of $n$ tokens, i.e., $ T = [ t_{1}, \ldots, t_{i-1}, t_{i}, \ldots, t_n ] $, when predicting a target token $t_{i}$, the generative language model takes the previous ones $\mathbf{t} = [t_{1},t_{2},...,t_{i-1}]$ as input, and then generates a probability distribution $\mathbf{p}$ over the vocabulary as output.   
In nearly all implementations, the Cross-Entropy loss is employed as the loss function, to maximize the predicted probability $\mathbf{p}_{t_i}$ w.r.t the ground-truth token $t_i$. Considering that the recent state-of-the-art deep language models (LM) predominantly leverage the Transformer architecture~\cite{DBLP:conf/nips/VaswaniSPUJGKP17}, the training loss $\mathcal{L}_{CE}$ of the generative language model can be formulated as follows.
\begin{align}
\label{eq:ce}
    \mathcal{L}_{CE} &= -\log(\mathbf{p}_{t_{i}}) \\
    s.t.,\quad\mathbf{p} &= \text{softmax}(W \mathbf{H}_{i-1}^{last}) \\
    \mathbf{H}^{last} &= \text{Transformer}(\text{Embedding}(\mathbf{t}))
\end{align}
Here, $\mathbf{H}^{last}$ denotes the hidden states of the last layer of the Transformer architecture, which consists of the hidden states w.r.t the previous tokens $\mathbf{t} = [t_{1},t_{2},...,t_{i-1}]$, i.e., $\mathbf{H}^{last}=[\mathbf{H}_{1}^{last},\mathbf{H}_{2}^{last}, \ldots, \mathbf{H}_{i-1}^{last}]$. With $\mathbf{H}^{last}_{i-1}$, a linear projection layer $W$ is introduced to derive the predicted probability distribution $\mathbf{p}$ over the vocabulary, with a \texttt{softmax} operation.


\paragraph{\textbf{Focal Loss for Classification}} 
Focal Loss is originally proposed for object detection to address the issue of extreme foreground-background class imbalance encountered during the training of one-stage object detectors~\cite{DBLP:conf/iccv/LinGGHD17}. 
Focal Loss can lead a classification model to concentrate more on a sparse set of difficult-to-learn classes and prevent the abundance of easy-to-learn classes from overwhelming the model during training. Actually, Focal Loss is an extension of Cross-Entropy Loss, with an extra dynamic scaling factor, as formulated below.
\begin{equation}
\begin{aligned}
    \mathcal{L}^0_{FL} &= -(1-p_{})^{\gamma }\log(p_{})
\end{aligned}
\end{equation}
Here, $p_{}$ is the predicted probability w.r.t the ground-truth class, and $\gamma$ is a hyperparameter with $\gamma \geq 0$. It can be seen that when $\gamma=0$, Focal Loss would degenerate to Cross-Entropy Loss. As $p$ decreases, i.e., getting more-difficult-to-learn, the dynamic scaling factor $(1-p)^{\gamma}$ increases, thus giving more attention (i.e., higher weights) to the difficult-to-learn classes.

\subsection{Focal Loss for Language Models} 
Generative language models are commonly trained on the massive textual corpus, which exhibits inherent token imbalance as revealed by Zipf's law~\cite{piantadosi2014zipf}. Such an imbalance of tokens can lead to two primary challenges:
1) Training efficiency becomes sub-optimal. A large number of easy-to-learn tokens (i.e., classes) provide marginal gains in learning signals.~\cite{DBLP:conf/iccv/LinGGHD17}. 
2) The training process can be overwhelmed by a large proportion of the frequent and easy-to-learn tokens, and thus pay insufficient attention to the other infrequent, difficult-to-learn but informative tokens, which might lead to performance degradation.

As revealed in Equation (\ref{eq:ce}), training a generative language model is essentially a classification problem. Therefore to mitigate the bias of learning difficulties caused by the inherent token imbalance, Focal Loss can be applied.  
Specifically, we can use the Focal Loss as a substitute for the Cross-Entropy Loss in Equation (\ref{eq:ce}) to train a generative language model as follows.
\begin{equation}
\begin{aligned}
    \mathcal{L}_{FL} &= -(1-\mathbf{p}_{t_{i}} )^{\gamma }\log(\mathbf{p}_{t_{i}})
\end{aligned}
\end{equation}
Here, the dynamic scaling factor $(1-\mathbf{p}_{t_{i}} )^{\gamma }$ is derived based on the predicted probability $\mathbf{p}_{t_{i}}$ of the to-be-learned token $t_{i}$.
Similarly, as the probability $\mathbf{p}_{t_{i}}$ decreases (i.e., being more difficult to learn), the scaling factor $(1-\mathbf{p}_{t_{i}} )^{\gamma }$ increases correspondingly. Therefore, more-difficult-to-learn tokens will receive higher loss weights. 

\subsection{Proposed MiLe Loss}

However, as illustrated in Figure \ref{fig:exaple} and analyzed before, in language model pretraining, predicting the next token is more like a multi-label classification problem.
When there are multiple valid next tokens for a given sequence of previous tokens, the learning difficulty assessed by Focal Loss is imperfect. 

To tackle that, we propose MiLe Loss, which leverages the information entropy of the predicted probability distribution $\mathbf{p}$ over the vocabulary, instead of the single probability $\mathbf{p}_{t_i}$ as Focal Loss, to derive a dynamic scaling factor. 
MiLe Loss is naturally designed for cases with multiple valid tokens.
It is inspired by the following observations: 1) when a next token is easy-to-learn, the minor valid tokens would divide up almost the total probability (i.e., 1.0) while others are associated with very low probabilities (i.e., $\mathbf{p}$ is more focused), resulting in a low information entropy; 2) when a next token is difficult-to-learn, the predicted probability distribution would be more uniform, resulting in a higher information entropy.


\begin{table*}[!ht]
\centering
\scalebox{0.9}{
\setlength{\tabcolsep}{4.8mm}{
\begin{tabular}{ccccccc}
\toprule
model size  & dimension & $n$ heads & $n$ layers & learning rate & batch size & seq length \\
\midrule
468M & 1024 & 16 & 24 & $3.0e^{-4}$ & 1024 & 1024\\
1.2B & 2048 & 8 & 16 & $3.0e^{-4}$ & 1024 & 1024\\
6.7B & 4096 & 32 & 32 & $3.0e^{-4}$ & 2048 & 2048\\
\bottomrule
\end{tabular}
}
}
\caption{Model sizes, architectures, and optimization hyper-parameters. }
\label{models}
\end{table*}

Specifically, MiLe Loss can be formulated as follows in language model pretraining.
\begin{equation}
\label{eq:ie}
\begin{aligned}
    \mathcal{L}_{IL} &= -(1- \sum_{j}\mathbf{p}_{j}\log(\mathbf{p}_{j})  )^{\gamma }\log(\mathbf{p}_{t_{i}})
\end{aligned}
\end{equation}
Here, $- \sum_{j}\mathbf{p}_{j}\log(\mathbf{p}_{j}) \geq 0$ is the information entropy of the predicted probability distribution $\mathbf{p}$ over the vocabulary. Note that when $\mathbf{p}$ is a uniform distribution, i.e., $p_j = \frac{1}{N}$ with $N$ being the vocabulary size for all $j$, the information entropy reaches its upper bound $\log(N)$. Therefore, the dynamic scaling factor $(1- \sum_{j}\mathbf{p}_{j}\log(\mathbf{p}_{j})  )$ is bounded in $[1, 1 + \log(N)]$.
When a next token is difficult to learn, the corresponding higher information entropy results in a higher scaling factor, and thus MiLe Loss increases the loss weights for such tokens. Conversely, MiLe Loss decreases the loss weights for easy-to-learn tokens, according to their lower information entropies.


\section{Experiments}
We train three generative language models of different capacities, i.e., 468M, 1.2B, and 6.7B parameters, on the open-source Pile dataset ~\cite{DBLP:journals/corr/abs-2101-00027} as ~\cite{DBLP:conf/icml/BidermanSABOHKP23,DBLP:journals/corr/abs-2305-10429,DBLP:conf/iclr/CarliniIJLTZ23}, and make comparisons among different loss functions. 

\subsection{The Pile dataset}
The Pile dataset is a public large-scale corpus for language model pretraining, which has over 825GB English texts across 22 domains. 
For experiments, we tokenize it using the remarkable LLaMA tokenizer ~\cite{DBLP:journals/corr/abs-2302-13971} with a 32k-sized vocabulary.
As the number of tokens changes with a new tokenizer, we follow~\cite{DBLP:journals/corr/abs-2305-10429} to re-calculate the sampling weight for each domain. Specifically, we chunk the dataset into sequences of 1,024 tokens, and then for each domain, we multiply its corresponding number of sequences with its domain-specific epochs reported in~\cite{DBLP:journals/corr/abs-2101-00027}. Finally, we normalize all the multiplication results to obtain the sampling weights listed in Table \ref{weights}.

\begin{table}[!t]
\centering
\scalebox{0.75}{
\begin{tabular}{lclc}
\toprule
  & Weights  &   & Weights  \\
\midrule
ArXiv & 0.1997 & OpenSubtitles & 0.0239 \\
BookCorpus2 & 0.0100 & OpenWebText2 & 0.1735 \\
Books3 & 0.1640 & PhilPapers & 0.0073 \\
DM Mathematics & 0.0502 & Pile-CC & 0.1551 \\
Enron Emails & 0.0030 & PubMed Abstracts & 0.0536 \\
EuroParl & 0.0156 & PubMed Central & 0.2823 \\
FreeLaw & 0.0895 & StackExchange & 0.1027 \\
Github & 0.0962 & USPTO Backgrounds & 0.0586 \\
Gutenberg(PG-19) & 0.0481 & Ubuntu IRC & 0.0229 \\
HackerNews & 0.0117 & Wikipedia(en) & 0.1121 \\
NIH ExPorter & 0.0047 & YoutubeSubtitles & 0.0151 \\
\bottomrule
\end{tabular}
}
\caption{Sampling weights on the Pile dataset.}
\label{weights}
\end{table}

\subsection{Experimental setup}
\begin{table*}[!t]
\centering
\scalebox{0.68}{
\begin{tabular}
{ll|p{1.1cm}<{\centering}p{1.65cm}<{\centering}p{1.65cm}<{\centering}p{1.65cm}<{\centering}p{1.65cm}<{\centering}p{1.65cm}<{\centering}p{1.65cm}<{\centering}p{1.7cm}<{\centering}|p{1.1cm}<{\centering}}
\toprule
 & & BoolQ & HellaSwag & LAMBADA & OpenBookQA & PIQA & SIQA & StoryCloze & Winogrande & Avg \\
 \midrule
 \multicolumn{9}{l}{\textit{468M}} \\
  \midrule
  \multirow{3}{*}{0-shot}
& Cross-Entropy Loss & 57.52 & 40.73 & 39.10 & 30.60 & 67.08 & 40.79 & 63.55 & 53.75 & 49.14 \\
& Focal Loss & 58.35 & 41.17 & 40.09 & \underline{\textbf{32.80}} & 67.25 & \underline{\textbf{41.91}} & 63.07 & 51.70 & 49.54 \\
& MiLe Loss & \underline{\textbf{59.57}} & \underline{\textbf{41.27}} & \underline{\textbf{41.34}} & 30.00 & \underline{\textbf{67.25}} & 41.61 & \underline{\textbf{63.60}} & \underline{\textbf{54.78}} & \underline{\textbf{49.93}} \\
\midrule
\multirow{3}{*}{1-shot}
& Cross-Entropy Loss & 54.22 & 40.86 & 37.16 & 30.40 & \underline{\textbf{67.85}} & 41.66 & 62.69 & 53.04 & 48.48 \\
& Focal Loss & 53.64 & \underline{\textbf{41.04}} & 37.88 & \underline{\textbf{32.20}} & 67.14 & \underline{\textbf{44.27}} & 62.16 & 52.64 & 48.87 \\
& MiLe Loss & \underline{\textbf{55.23}} & 40.90 & \underline{\textbf{38.75}} & 32.00 & 67.68 & 43.35 & \underline{\textbf{63.23}} & \underline{\textbf{55.88}} & \underline{\textbf{49.63}} \\
\midrule
\multirow{3}{*}{5-shot}
& Cross-Entropy Loss & 50.89 & 41.06 & 36.27 & 28.80 & \underline{\textbf{67.68}} & 43.39 & 62.37 & 50.99 & 47.68 \\
& Focal Loss & 48.10 & \underline{\textbf{41.80}} & 38.50 & \underline{\textbf{31.40}} & 67.19 & \underline{\textbf{46.01}} & \underline{\textbf{63.01}} & 52.09 & 48.51 \\
& MiLe Loss & \underline{\textbf{52.29}} & 41.53 & \underline{\textbf{39.05}} & 28.80 & 67.41 & 45.39 & 62.85 & \underline{\textbf{54.06}} & \underline{\textbf{48.92}} \\
\midrule
 \multicolumn{9}{l}{\textit{1.2B}} \\
  \midrule
  \multirow{3}{*}{0-shot}
& Cross-Entropy Loss & 55.96 & 47.48 & 45.76 & 32.20 & 69.64 & 42.43 & 65.47 & 54.54 & 51.69 \\
& Focal Loss & \underline{\textbf{62.02}} & 47.61 & 46.87 & 33.00 & 69.59 & \underline{\textbf{42.02}} & 65.63 & 55.01 & \underline{\textbf{52.72}} \\
& MiLe Loss & 56.94 & \underline{\textbf{47.64}} & \underline{\textbf{47.37}} & \underline{\textbf{33.80}} & \underline{\textbf{70.13}} & 41.91 & \underline{\textbf{66.06}} & \underline{\textbf{55.96}} & 52.48 \\
\midrule
 \multirow{3}{*}{1-shot}
& Cross-Entropy Loss & 54.71 &47.37 &42.13 &34.40 &69.42 &44.78 &65.26 &\underline{\textbf{56.27}} &51.79 \\
& Focal Loss & \underline{\textbf{62.35}} &\underline{\textbf{47.41}} &43.88 &32.60 &69.15 &45.04 &65.42 &54.85 &\underline{\textbf{52.59}} \\
& MiLe Loss & 54.95 &47.39 &\underline{\textbf{45.08}} &\underline{\textbf{34.00}} &\underline{\textbf{70.13}} &\underline{\textbf{45.04}} &\underline{\textbf{65.58}} &54.85 &52.13 \\
\midrule
\multirow{3}{*}{5-shot}
& Cross-Entropy Loss & 55.72 &47.74 &41.55 &33.00 &69.86 &45.04 &66.11 &55.64 &51.83 \\
& Focal Loss & \underline{\textbf{62.17}} &\underline{\textbf{48.00}} &42.87 &32.00 &69.75 &45.60 &66.01 &56.20 &\underline{\textbf{52.82}} \\
& MiLe Loss & 55.38 &47.78 &\underline{\textbf{45.00}} &\underline{\textbf{34.00}} &\underline{\textbf{70.13}} &\underline{\textbf{46.26}} &\underline{\textbf{66.22}} &\underline{\textbf{56.83}} &52.70 \\

\midrule
\multicolumn{9}{l}{\textit{6.7B}} \\
 \midrule
  \multirow{3}{*}{0-shot}
& Cross-Entropy Loss & \underline{\textbf{62.14}} & 58.91 & 55.54 & 34.40 & 73.61 & 44.06 & 70.66 & 61.40 & 57.59 \\
& Focal Loss & 59.72 & 59.59 & 55.64 & \underline{\textbf{36.60}} & 73.94 & 43.04 & 70.12 & \underline{\textbf{61.88}} & 57.57 \\
& MiLe Loss & 60.89 & \underline{\textbf{59.63}} & \underline{\textbf{57.73}} & 35.20 & \underline{\textbf{73.99}} & \underline{\textbf{44.06}} & \underline{\textbf{71.25}} & 61.01 & \underline{\textbf{57.97}}\\
\midrule
 \multirow{3}{*}{1-shot}
& Cross-Entropy Loss & 59.24 & 58.68 & 53.48 & 37.00 & 73.99 & 47.90 & 70.60 & 60.69 & 57.70 \\
& Focal Loss & 58.53 & 59.23 & 52.59 & 35.60 & \underline{\textbf{74.27}} & 48.06 & 69.96 & 59.91 & 57.27 \\
& MiLe Loss & \underline{\textbf{60.46}} & \underline{\textbf{59.56}} & \underline{\textbf{55.35}} & \underline{\textbf{38.00}} & 73.29 & \underline{\textbf{48.57}} & \underline{\textbf{70.87}} & \underline{\textbf{61.01}} & \underline{\textbf{58.39}} \\
\midrule
\multirow{3}{*}{5-shot}
& Cross-Entropy Loss & 61.28 & 59.44 & 54.01 & 37.00 & \underline{\textbf{74.16}} & 49.03 & 71.30 & 63.06 & 58.66 \\
& Focal Loss & 57.98 & \underline{\textbf{60.10}} & 55.91 & 36.80 & 74.05 & 50.0 & 70.44 & 62.90 & 58.52 \\
& MiLe Loss & \underline{\textbf{62.20}} & 60.06 & \underline{\textbf{58.16}} & \underline{\textbf{37.80}} & 73.61 & \underline{\textbf{50.67}} & \underline{\textbf{71.67}} & \underline{\textbf{63.30}} & \underline{\textbf{59.68}} \\
\midrule
\end{tabular}
}
\caption{Zero-shot and few-shot performance (i.e., accuracy) of models at different scales on common sense reasoning benchmarks.}
\label{expert_results}
\end{table*}

We train three generative language models with 468M, 1.2B, and 6.7B parameters, respectively. Specifically, the architectures of the 468M-parameter and the 1.2B-parameter models, including the dimensionality of hidden states, the number of layers, etc., are identical to those of the 410M-parameter and the 1.0B-parameter models outlined in \cite{DBLP:conf/icml/BidermanSABOHKP23}. The minor differences in parameter sizes are attributed to the variations of vocabulary size in the embedding layer. As for the 6.7B-parameter model, its architecture is identical to LLaMA-7B~\cite{DBLP:journals/corr/abs-2302-13971}. The corresponding hyperparameters for each model can be found in Table \ref{models}.
Following LLaMA~\cite{DBLP:journals/corr/abs-2302-13971}, we use the AdamW optimizer~\cite{DBLP:conf/iclr/LoshchilovH19} with a learning rate of $3.0e^{-4}$, $2k$ warmup steps, and a cosine learning rate decay schedule. 
Following ~\cite{DBLP:conf/iccv/LinGGHD17}, the hyperparameter $\gamma$ is set as $1.0$ for both Focal Loss and the proposed MiLe Loss, unless explicitly stated otherwise. 
Due to the computational budget and following the pretraining settings of~\cite{DBLP:journals/corr/abs-2305-10429}, all models are pretrained with 100B tokens. 

Following~\cite{DBLP:journals/corr/abs-2302-13971,DBLP:conf/nips/BrownMRSKDNSSAA20,DBLP:journals/corr/abs-2112-11446,DBLP:journals/corr/abs-2203-15556}, we primarily evaluate all models on tasks of common-sense reasoning, closed-book question answering, and massive multitask language understanding. For fair comparisons, we utilize the open-source pipeline \texttt{lm-evaluation-harness}\footnote{https://github.com/EleutherAI/lm-evaluation-harness}~\cite{eval-harness} for evaluation, as~\cite{DBLP:conf/icml/BidermanSABOHKP23,DBLP:conf/icml/DettmersZ23}.

\subsection{Experimental Results}
\paragraph{\textbf{Common Sense Reasoning}}

Following~\cite{DBLP:journals/corr/abs-2302-13971,DBLP:conf/nips/BrownMRSKDNSSAA20,DBLP:journals/corr/abs-2112-11446,DBLP:journals/corr/abs-2203-15556}, we employ 8 widely used benchmark datasets for the evaluation of common sense reasoning, including BoolQ~\cite{DBLP:conf/naacl/ClarkLCK0T19}, HellaSwag~\cite{DBLP:conf/acl/ZellersHBFC19}, LAMBADA~\cite{DBLP:conf/acl/PapernoKLPBPBBF16}, OpenBookQA~\cite{DBLP:conf/emnlp/MihaylovCKS18}, PIQA~\cite{DBLP:conf/aaai/BiskZLGC20}, SIQA~\cite{DBLP:journals/corr/abs-1904-09728}, StoryCloze~\cite{DBLP:journals/corr/MostafazadehCHP16},Winogrande~\cite{DBLP:conf/aaai/SakaguchiBBC20}. We report the model performance in terms of accuracy for zero-shot and few-shot settings in Table \ref{expert_results}, like~\cite{DBLP:journals/corr/abs-2302-13971,DBLP:conf/nips/BrownMRSKDNSSAA20}. 

We can observe that the proposed MiLe Loss substantially outperforms both Cross-Entropy Loss and Focal Loss on different setups with different model capacities. Specifically, for models with 468M and 6.7B parameters on 0/1/5-shot settings, MiLe Loss consistently achieves superior performance to both compared baselines. As for the 1.2B-parameter model, although MiLe Loss yields slightly lower average performance than Focal Loss, it still delivers the highest performance on 6 out of the 8 datasets and steadily outperforms Cross-Entropy Loss on most datasets.

These results clearly demonstrate the effectiveness of the proposed MiLe Loss.
We attribute it to that MiLe Loss compels language models to allocate more attention to those difficult-to-learn yet informative tokens during pretraining, which mitigates the bias of learning difficulties among tokens. 
Moreover, the consistent performance superiority of MiLe Loss over Focal Loss also validates that, relying on the information entropy of the predicted probability distribution over the vocabulary to assess the learning difficulties of tokens is more reasonable.

\paragraph{\textbf{Closed Book Question Answering}}

\begin{table}[!t]
\centering
\scalebox{0.95}{
\begin{tabular}{l|cccc}
\toprule
& 0-shot & 1-shot & 5-shot  \\
\midrule
\multicolumn{4}{l}{\textit{TriviaQA}} \\
\midrule
 Cross-Entropy Loss & 17.09 & 21.98 & 26.33 \\
 Focal Loss & 16.47 & 23.03  & 27.31 \\
MiLe Loss & \textbf{20.64} & \textbf{23.42}  & \textbf{28.75} \\
\midrule
\multicolumn{4}{l}{\textit{WebQuestions}} \\
\midrule
Cross-Entropy Loss & \textbf{5.22} & 9.79 & 14.17 \\
Focal Loss & 4.53 & 9.60 & \textbf{14.62} \\
MiLe Loss & 5.02 & \textbf{9.89} & 14.57 \\
\midrule
\end{tabular}
}
\caption{Zero-shot and few-shot exact match performance of 6.7B-parameter models on closed-book question-answering benchmarks.}
\label{closed-book}
\end{table}

Following~\cite{DBLP:conf/nips/BrownMRSKDNSSAA20,DBLP:journals/corr/abs-2302-13971}, for the task of closed book question answering, we evaluate the performance of the largest 6.7B-parameter models with different loss functions on two benchmark datasets, i.e., TriviaQA~\cite{DBLP:conf/acl/JoshiCWZ17} and WebQuestions~\cite{DBLP:conf/emnlp/BerantCFL13}. We report the exact match performance for the zero-shot and few-shot settings in Table \ref{closed-book}.

It can be seen that language models trained with the proposed MiLe Loss achieve superior performance across most settings. Compared with Cross-Entropy Loss, MiLe Loss achieves substantial performance improvement in 5 out of 6 settings. 
Particularly, on TriviaQA, MiLe Loss achieves a maximum performance improvement of 3.55\% (0-shot) over Cross-Entropy Loss.
Compared with Focal Loss, MiLe Loss also exhibits consistent superiority. Notably, in the 0-shot setting on TriviaQA, MiLe Loss outperforms Focal Loss by 4.17\%. 

\paragraph{\textbf{Massive Multitask Language Understanding}}
We further validate the effectiveness of the proposed MiLe Loss on the MMLU (Massive Multitask Language Understanding) benchmark~\cite{DBLP:conf/iclr/HendrycksBBZMSS21}. MMLU  consists of multiple-choice questions covering 57 subjects, 
including STEM, social sciences, humanities, etc.
It has been serving as a benchmark for evaluating the multitasking capability of pretrained language models.
Following LLaMA~\cite{DBLP:journals/corr/abs-2302-13971}, we evaluate the 6.7B-parameter models in the 5-shot setting. Among multiple choices, we choose the one with the highest probability normalized by the number of tokens. 

As shown in Table \ref{mmlu}, MiLe loss exhibits superior performance on average. Compared with Cross-Entropy loss, MiLe loss obtains performance improvement of 0.32\%, 1.28\%, and 0.91\% for the field of \texttt{STEM}, \texttt{Humanities}, and \texttt{Other}, respectively. For the field of \texttt{Social Sciences}, the performance decline may be attributed to that MiLe Loss tends to consider \texttt{Social Sciences} samples as easier-to-learn ones. We intend to study it in depth in our future work. Compared with Focal Loss, MiLe Loss also yields superior performance on all fields except \texttt{STEM}. All the results above further demonstrate the proposed MiLe Loss's effectiveness and reasonableness.

\section{Analyses}
We conduct further experiments to provide more insightful analyses on the proposed MiLe Loss. 

\subsection{Impact of $\gamma$}


\begin{table}[!t]
\centering
\scalebox{0.85}{
\begin{tabular}{l|ccc}
\toprule
 &  Cross-Entropy &Focal& MiLe \\
 &  Loss &Loss& Loss \\
\midrule
STEM & 29.59 &\textbf{29.99} &29.91 \\
Social Sciences & \textbf{29.64} &27.57 &28.07\\
Humanities & 27.00 &27.35 &\textbf{28.28}\\
Other & 29.94 &29.34 &\textbf{30.85} \\
\midrule
Avg & 29.38 &28.90 &\textbf{29.68}\\
\midrule
\end{tabular}
}
\caption{The 5-shot learning performance of 6.7B-parameter models on MMLU.}
\label{mmlu}
\end{table}

\begin{figure*}[!t]
    \centering
    \includegraphics[width=1\textwidth]{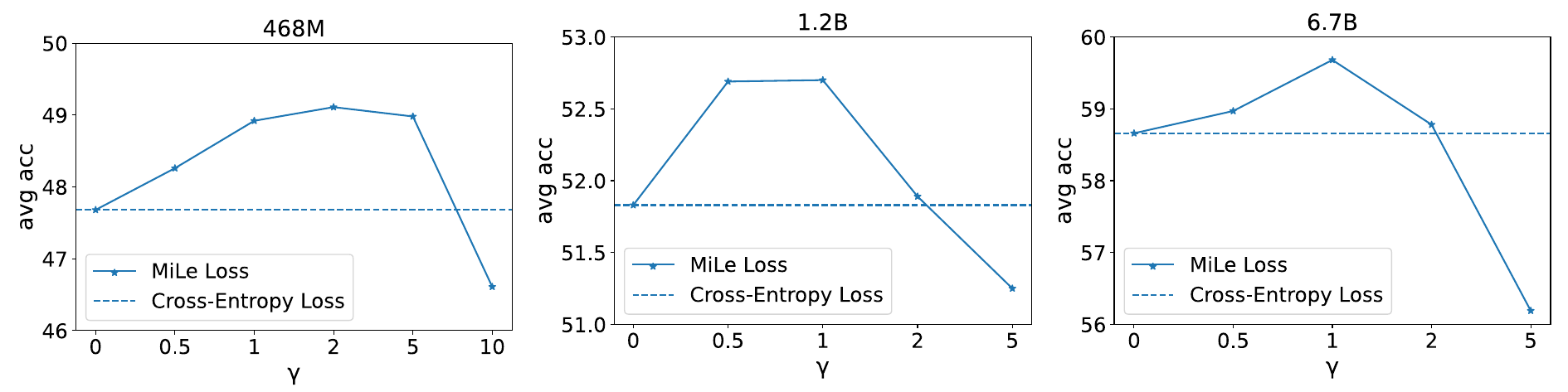}
    \caption{The performance of MiLe Loss and Cross-Entropy Loss in 5-shot learning with different $\gamma$ values.}
    \label{fig:focal_entropy_compare}
\end{figure*}

We aim to discern the performance change of the proposed MiLe Loss on language models with different values of $\gamma$, i.e., the hyperparameter in Equation (\ref{eq:ie}). It's worth noting that when $\gamma$ is set to 0, MiLe Loss is functionally equivalent to Cross-Entropy Loss. 
As $\gamma$ increases, the language model becomes more focused on the difficult-to-learn tokens, i.e., those with higher information entropy. 
Here we conduct a grid search for $\gamma$ on language models of various scales (i.e., 468M, 1.2B, and 6.7B parameters), and use the average performance in 5-shot learning for the Common Sense Reasoning task that covers the most benchmarks as the evaluation metric.

As shown in Figure \ref{fig:focal_entropy_compare}, when $\gamma$ increases from $0$ to $5$ for the 468M-parameter model or increases from $0$ to $2$ for the 1.2B-parameter/6.7B-parameter models, the performances of MiLe Loss consistently surpass those of Cross-Entropy Loss. The results clearly demonstrate that the performance of MiLe Loss is not very sensitive to the setting of the hyperparameter $\gamma$, which shows practical applicability. As expected, when $\gamma$ increases to a relatively large value, the performance of MiLe Loss declines, because too much attention is given to the difficult-to-learn tokens, and the easy-to-learn ones get overlooked as a result.

\subsection{Perplexity on the Pile Validation Set}
Here we further discuss how the proposed MiLe Loss affects the perplexity of pretrained language models on the Pile validation set. 

Table \ref{validation_ppl} reports the perplexity of the largest 6.7B-parameter models trained with $\gamma$ increasing from $0$ to $5$ for MiLe Loss. Among them, $\gamma = 0$ is equivalent to Cross-Entropy Loss. Notably, when $\gamma=0.5$, the perplexity obtained by MiLe Loss is lower than that by Cross-Entropy Loss (i.e., $\gamma=0$). However, as we increase $\gamma$, the perplexity of MiLe Loss also increases and becomes higher than that of Cross-Entropy Loss. The increase of perplexity can be attributed to: 1) the measurement of perplexity is directly related to the exponentiation of Cross-Entropy Loss, and thus optimizing Cross-Entropy Loss during training is consistent with optimizing the perplexity; 2) the objective function of MiLe Loss somewhat diverges from that of perplexity due to the dynamic scaling factor, and thus optimizing it may lead to an increase of perplexity.

\begin{figure}[!t]
    \centering
    \includegraphics[width=0.5\textwidth]{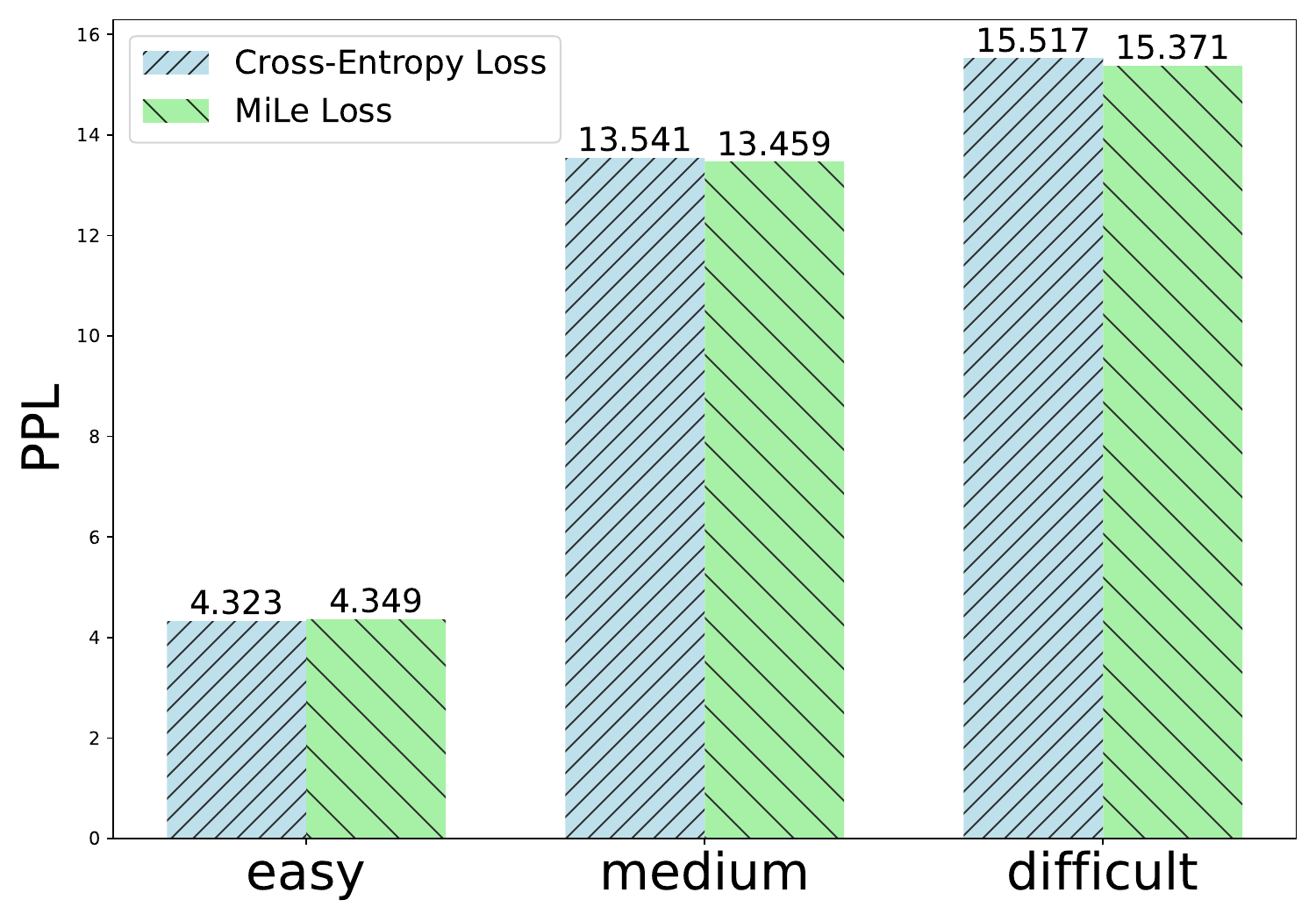}
    \caption{The average perplexity (i.e., PPL) for tokens in different learning-difficulty levels.}
    \label{difficulty}
\end{figure}

To thoroughly inspect how the perplexity increases, we conduct a fine-grained analysis of perplexity at the token level.
Similar to the perplexity analysis before, we group all tokens into three learning-difficulty levels based on their corresponding frequencies, i.e., \texttt{easy}, \texttt{medium}, and \texttt{difficult}. 
Specifically, we categorize the top tokens that cover $80\%$ of the Pile dataset as \texttt{easy}, those that cover the extra $15\%$ (i.e., $80\%-95\%$) of the Pile dataset as \texttt{medium}, and the remaining $5\%$ as \texttt{difficult}.
The average perplexity for tokens in each learning-difficulty level, obtained by Cross-Entropy Loss and the proposed MiLe Loss with $\gamma=1$, is shown in Figure \ref{difficulty}. It can be seen that, compared with Cross-Entropy Loss, MiLe Loss results in an unnoticeable increase in perplexity for the \texttt{easy} tokens, while for the \texttt{medium} or the \texttt{difficult} tokens, MiLe Loss substantially reduces their perplexity with a noticeable decline. Given that \texttt{easy} tokens dominate the dataset, the overall increase in perplexity is expected. However, the substantial decline of perplexity for the \texttt{medium} or the \texttt{difficult} tokens further demonstrates the effectiveness of MiLe Loss in guiding language models to focus more on infrequent, difficult-to-learn but informative tokens and thereby mitigating the bias of learning difficulties during training.
\begin{table}[!t]
\centering
\scalebox{0.9}{
\begin{tabular}{l|ccccc}
\toprule
 $\gamma$ &  0 &  0.5 & 1 & 2 & 5  \\
\midrule
PPL & 5.473 & 5.467 & 5.492 &5.608 &6.317 \\
\bottomrule
\end{tabular}
}
\caption{The perplexity (PPL) on the Pile validation set under different $\gamma$ values for MiLe Loss. Among them, $\gamma=0$ equals Cross-Entropy Loss.}
\label{validation_ppl}
\end{table}

\begin{table*}[!t]
\centering
\scalebox{0.68}{
\begin{tabular}
{ll|p{1.1cm}<{\centering}p{1.65cm}<{\centering}p{1.65cm}<{\centering}p{1.65cm}<{\centering}p{1.65cm}<{\centering}p{1.65cm}<{\centering}p{1.65cm}<{\centering}p{1.7cm}<{\centering}|p{1.1cm}<{\centering}}
\toprule
 & & BoolQ & HellaSwag & LAMBADA & OpenBookQA & PIQA & SIQA & StoryCloze & Winogrande & Avg \\
\midrule
  \multirow{2}{*}{0-shot}
& Cross-Entropy Loss & 66.73 & 63.48 & 60.95 & 36.80 & \underline{\textbf{75.52}} & 44.58 & 72.26 & 61.88 & 60.27 \\
& MiLe Loss & \underline{\textbf{68.62}} & \underline{\textbf{64.17}} & \underline{\textbf{61.52}} & \underline{\textbf{39.00}} & 75.41 & \underline{\textbf{44.63}} & \underline{\textbf{72.90}} & \underline{\textbf{63.61}} & \underline{\textbf{61.23}} \\
\midrule
 \multirow{2}{*}{1-shot}
& Cross-Entropy Loss & 64.13 & 63.33 & 58.92 & \underline{\textbf{40.40}} & 75.46 & 48.72 & \underline{\textbf{72.90}} & 63.85 & 60.96 \\
& MiLe Loss & \underline{\textbf{65.26}} & \underline{\textbf{63.93}} & \underline{\textbf{60.57}} & 38.80 & \underline{\textbf{75.46}} & \underline{\textbf{49.64}} & 72.58 & \underline{\textbf{63.93}} & \underline{\textbf{61.27}} \\
\midrule
\multirow{2}{*}{5-shot}
& Cross-Entropy Loss & 64.22 & 63.92 & 60.90 & 39.60 & \underline{\textbf{75.84}} & 51.18 & 73.60 & 64.72 & 61.75 \\
& MiLe Loss & \underline{\textbf{66.85}} & \underline{\textbf{64.58}} & \underline{\textbf{64.33}} & \underline{\textbf{41.00}} & 75.14 & \underline{\textbf{52.66}} & \underline{\textbf{74.02}} & \underline{\textbf{66.06}} & \underline{\textbf{63.08}} \\
\midrule
\end{tabular}
}
\caption{The performance of the 6.7B models trained with 200B tokens in zero/few-shot settings across various benchmarks.}
\label{continue_pretraining}
\end{table*}

\subsection{Performance of Training with More Tokens}

MiLe Loss assesses the learning difficulty of each token through information entropy. Intuitively, the more tokens used in model training, the more powerful the Language Model becomes, and the output word distribution becomes more reasonable. Consequently, the assessment of the learning difficulties of tokens becomes more accurate, and thus the MiLe Loss can probably better lead the LM to tackle the bias. To validate that, with limited computational resources, we continue to pre-train the 6.7B model from 100B tokens to 200B tokens, with both Cross-Entropy Loss and MiLe Loss. Their corresponding evaluation results on all benchmarks are reported in Table \ref{continue_pretraining}. We can see that, when the number of training tokens for the 6.7B models increases to 200B, the models trained with MiLe Loss yield consistent and substantial performance improvements over those trained with Cross-Entropy Loss. Moreover, compared to training with 100B tokens, training with more tokens even helps MiLe Loss to yield \textbf{LARGER} performance improvements. For instance, in the 5-shot setting, with 100B training tokens, the performance improvements gained by MiLe Loss over Cross-Entropy Loss on the 6.7B models is 1.02\%. Then by continuing pre-training with more training tokens, the gained improvements increase to 1.33\%. The experimental results above demonstrate well that using more tokens increases the benefits of MiLe Loss.

\section{Conclusions}

In this paper, we present our observation of the bias of learning difficulties among tokens during language model pretraining, mainly caused by the inherent token imbalance in textual training data. 
We initially introduce Focal Loss as an attempt to mitigate the bias of learning difficulties. However, we find that considering the single probability of the ground-truth next token for assessing its learning difficulty is unreasonable, especially in cases with multiple valid next tokens. 
To tackle that, we propose MiLe Loss, which assesses the learning difficulty of a token by taking into account the global information entropy of the predicted probability distribution over the vocabulary. Extensive experiments demonstrate that, compared with both Cross-Entropy Loss and Focal Loss, the proposed MiLe Loss achieves superior performance for various downstream tasks in zero-shot and few-shot learning settings.

\section{Limitations}

In the proposed MiLe Loss, we scale the Cross-Entropy Loss based on information entropy to lead a generative language model to allocate more attention to difficult-to-learn tokens, which yields superior performance. 
Yet the effectiveness of MiLe Loss may be influenced by the quality of the training data.
Specifically, as noisy data samples are generally outliers, the predicted probability distributions on them would typically exhibit high information entropy. Thus, too many noisy samples may make MiLe Loss amplify their corresponding loss weights too much, causing negative impacts on the model performance. 
We leave the investigation of how noisy data samples affect MiLe Loss to our future research.
\bibliography{anthology,custom}

\begin{thebibliography}{50}
\expandafter\ifx\csname natexlab\endcsname\relax\def\natexlab#1{#1}\fi

\bibitem[{Ando and Huang(2017)}]{DBLP:conf/pkdd/AndoH17}
Shin Ando and Chun{-}Yuan Huang. 2017.
\newblock \href {https://doi.org/10.1007/978-3-319-71249-9\_46} {Deep over-sampling framework for classifying imbalanced data}.
\newblock In \emph{Machine Learning and Knowledge Discovery in Databases - European Conference, {ECML} {PKDD} 2017, Skopje, Macedonia, September 18-22, 2017, Proceedings, Part {I}}, volume 10534 of \emph{Lecture Notes in Computer Science}, pages 770--785. Springer.

\bibitem[{Bahl et~al.(1983)Bahl, Jelinek, and Mercer}]{DBLP:journals/pami/BahlJM83}
Lalit~R. Bahl, Frederick Jelinek, and Robert~L. Mercer. 1983.
\newblock \href {https://doi.org/10.1109/TPAMI.1983.4767370} {A maximum likelihood approach to continuous speech recognition}.
\newblock \emph{{IEEE} Trans. Pattern Anal. Mach. Intell.}, 5(2):179--190.

\bibitem[{Berant et~al.(2013)Berant, Chou, Frostig, and Liang}]{DBLP:conf/emnlp/BerantCFL13}
Jonathan Berant, Andrew Chou, Roy Frostig, and Percy Liang. 2013.
\newblock \href {https://aclanthology.org/D13-1160/} {Semantic parsing on freebase from question-answer pairs}.
\newblock In \emph{Proceedings of the 2013 Conference on Empirical Methods in Natural Language Processing, {EMNLP} 2013, 18-21 October 2013, Grand Hyatt Seattle, Seattle, Washington, USA, {A} meeting of SIGDAT, a Special Interest Group of the {ACL}}, pages 1533--1544. {ACL}.

\bibitem[{Biderman et~al.(2023)Biderman, Schoelkopf, Anthony, Bradley, O'Brien, Hallahan, Khan, Purohit, Prashanth, Raff, Skowron, Sutawika, and van~der Wal}]{DBLP:conf/icml/BidermanSABOHKP23}
Stella Biderman, Hailey Schoelkopf, Quentin~Gregory Anthony, Herbie Bradley, Kyle O'Brien, Eric Hallahan, Mohammad~Aflah Khan, Shivanshu Purohit, USVSN~Sai Prashanth, Edward Raff, Aviya Skowron, Lintang Sutawika, and Oskar van~der Wal. 2023.
\newblock \href {https://proceedings.mlr.press/v202/biderman23a.html} {Pythia: {A} suite for analyzing large language models across training and scaling}.
\newblock In \emph{International Conference on Machine Learning, {ICML} 2023, 23-29 July 2023, Honolulu, Hawaii, {USA}}, volume 202 of \emph{Proceedings of Machine Learning Research}, pages 2397--2430. {PMLR}.

\bibitem[{Bisk et~al.(2020)Bisk, Zellers, Bras, Gao, and Choi}]{DBLP:conf/aaai/BiskZLGC20}
Yonatan Bisk, Rowan Zellers, Ronan~Le Bras, Jianfeng Gao, and Yejin Choi. 2020.
\newblock \href {https://doi.org/10.1609/aaai.v34i05.6239} {{PIQA:} reasoning about physical commonsense in natural language}.
\newblock In \emph{The Thirty-Fourth {AAAI} Conference on Artificial Intelligence, {AAAI} 2020, The Thirty-Second Innovative Applications of Artificial Intelligence Conference, {IAAI} 2020, The Tenth {AAAI} Symposium on Educational Advances in Artificial Intelligence, {EAAI} 2020, New York, NY, USA, February 7-12, 2020}, pages 7432--7439. {AAAI} Press.

\bibitem[{Brown et~al.(2020)Brown, Mann, Ryder, Subbiah, Kaplan, Dhariwal, Neelakantan, Shyam, Sastry, Askell, Agarwal, Herbert{-}Voss, Krueger, Henighan, Child, Ramesh, Ziegler, Wu, Winter, Hesse, Chen, Sigler, Litwin, Gray, Chess, Clark, Berner, McCandlish, Radford, Sutskever, and Amodei}]{DBLP:conf/nips/BrownMRSKDNSSAA20}
Tom~B. Brown, Benjamin Mann, Nick Ryder, Melanie Subbiah, Jared Kaplan, Prafulla Dhariwal, Arvind Neelakantan, Pranav Shyam, Girish Sastry, Amanda Askell, Sandhini Agarwal, Ariel Herbert{-}Voss, Gretchen Krueger, Tom Henighan, Rewon Child, Aditya Ramesh, Daniel~M. Ziegler, Jeffrey Wu, Clemens Winter, Christopher Hesse, Mark Chen, Eric Sigler, Mateusz Litwin, Scott Gray, Benjamin Chess, Jack Clark, Christopher Berner, Sam McCandlish, Alec Radford, Ilya Sutskever, and Dario Amodei. 2020.
\newblock \href {https://proceedings.neurips.cc/paper/2020/hash/1457c0d6bfcb4967418bfb8ac142f64a-Abstract.html} {Language models are few-shot learners}.
\newblock In \emph{Advances in Neural Information Processing Systems 33: Annual Conference on Neural Information Processing Systems 2020, NeurIPS 2020, December 6-12, 2020, virtual}.

\bibitem[{Buda et~al.(2018)Buda, Maki, and Mazurowski}]{DBLP:journals/nn/BudaMM18}
Mateusz Buda, Atsuto Maki, and Maciej~A. Mazurowski. 2018.
\newblock \href {https://doi.org/10.1016/j.neunet.2018.07.011} {A systematic study of the class imbalance problem in convolutional neural networks}.
\newblock \emph{Neural Networks}, 106:249--259.

\bibitem[{Carlini et~al.(2023)Carlini, Ippolito, Jagielski, Lee, Tram{\`{e}}r, and Zhang}]{DBLP:conf/iclr/CarliniIJLTZ23}
Nicholas Carlini, Daphne Ippolito, Matthew Jagielski, Katherine Lee, Florian Tram{\`{e}}r, and Chiyuan Zhang. 2023.
\newblock \href {https://openreview.net/pdf?id=TatRHT\_1cK} {Quantifying memorization across neural language models}.
\newblock In \emph{The Eleventh International Conference on Learning Representations, {ICLR} 2023, Kigali, Rwanda, May 1-5, 2023}. OpenReview.net.

\bibitem[{Chawla et~al.(2002)Chawla, Bowyer, Hall, and Kegelmeyer}]{DBLP:journals/jair/ChawlaBHK02}
Nitesh~V. Chawla, Kevin~W. Bowyer, Lawrence~O. Hall, and W.~Philip Kegelmeyer. 2002.
\newblock \href {https://doi.org/10.1613/jair.953} {{SMOTE:} synthetic minority over-sampling technique}.
\newblock \emph{J. Artif. Intell. Res.}, 16:321--357.

\bibitem[{Chen et~al.(2018)Chen, Ding, Lin, Zhao, and Han}]{DBLP:conf/ijcai/ChenDLZH18}
Hui Chen, Guiguang Ding, Zijia Lin, Sicheng Zhao, and Jungong Han. 2018.
\newblock \href {https://doi.org/10.24963/IJCAI.2018/84} {Show, observe and tell: Attribute-driven attention model for image captioning}.
\newblock In \emph{Proceedings of the Twenty-Seventh International Joint Conference on Artificial Intelligence, {IJCAI} 2018, July 13-19, 2018, Stockholm, Sweden}, pages 606--612. ijcai.org.

\bibitem[{Chowdhery et~al.(2022)Chowdhery, Narang, Devlin, Bosma, Mishra, Roberts, Barham, Chung, Sutton, Gehrmann, Schuh, Shi, Tsvyashchenko, Maynez, Rao, Barnes, Tay, Shazeer, Prabhakaran, Reif, Du, Hutchinson, Pope, Bradbury, Austin, Isard, Gur{-}Ari, Yin, Duke, Levskaya, Ghemawat, Dev, Michalewski, Garcia, Misra, Robinson, Fedus, Zhou, Ippolito, Luan, Lim, Zoph, Spiridonov, Sepassi, Dohan, Agrawal, Omernick, Dai, Pillai, Pellat, Lewkowycz, Moreira, Child, Polozov, Lee, Zhou, Wang, Saeta, Diaz, Firat, Catasta, Wei, Meier{-}Hellstern, Eck, Dean, Petrov, and Fiedel}]{DBLP:journals/corr/abs-2204-02311}
Aakanksha Chowdhery, Sharan Narang, Jacob Devlin, Maarten Bosma, Gaurav Mishra, Adam Roberts, Paul Barham, Hyung~Won Chung, Charles Sutton, Sebastian Gehrmann, Parker Schuh, Kensen Shi, Sasha Tsvyashchenko, Joshua Maynez, Abhishek Rao, Parker Barnes, Yi~Tay, Noam Shazeer, Vinodkumar Prabhakaran, Emily Reif, Nan Du, Ben Hutchinson, Reiner Pope, James Bradbury, Jacob Austin, Michael Isard, Guy Gur{-}Ari, Pengcheng Yin, Toju Duke, Anselm Levskaya, Sanjay Ghemawat, Sunipa Dev, Henryk Michalewski, Xavier Garcia, Vedant Misra, Kevin Robinson, Liam Fedus, Denny Zhou, Daphne Ippolito, David Luan, Hyeontaek Lim, Barret Zoph, Alexander Spiridonov, Ryan Sepassi, David Dohan, Shivani Agrawal, Mark Omernick, Andrew~M. Dai, Thanumalayan~Sankaranarayana Pillai, Marie Pellat, Aitor Lewkowycz, Erica Moreira, Rewon Child, Oleksandr Polozov, Katherine Lee, Zongwei Zhou, Xuezhi Wang, Brennan Saeta, Mark Diaz, Orhan Firat, Michele Catasta, Jason Wei, Kathy Meier{-}Hellstern, Douglas Eck, Jeff Dean, Slav Petrov, and Noah Fiedel.
  2022.
\newblock \href {https://doi.org/10.48550/arXiv.2204.02311} {Palm: Scaling language modeling with pathways}.
\newblock \emph{CoRR}, abs/2204.02311.

\bibitem[{Clark et~al.(2019)Clark, Lee, Chang, Kwiatkowski, Collins, and Toutanova}]{DBLP:conf/naacl/ClarkLCK0T19}
Christopher Clark, Kenton Lee, Ming{-}Wei Chang, Tom Kwiatkowski, Michael Collins, and Kristina Toutanova. 2019.
\newblock \href {https://doi.org/10.18653/v1/n19-1300} {Boolq: Exploring the surprising difficulty of natural yes/no questions}.
\newblock In \emph{Proceedings of the 2019 Conference of the North American Chapter of the Association for Computational Linguistics: Human Language Technologies, {NAACL-HLT} 2019, Minneapolis, MN, USA, June 2-7, 2019, Volume 1 (Long and Short Papers)}, pages 2924--2936. Association for Computational Linguistics.

\bibitem[{Cui et~al.(2019)Cui, Jia, Lin, Song, and Belongie}]{DBLP:conf/cvpr/CuiJLSB19}
Yin Cui, Menglin Jia, Tsung{-}Yi Lin, Yang Song, and Serge~J. Belongie. 2019.
\newblock \href {https://doi.org/10.1109/CVPR.2019.00949} {Class-balanced loss based on effective number of samples}.
\newblock In \emph{{IEEE} Conference on Computer Vision and Pattern Recognition, {CVPR} 2019, Long Beach, CA, USA, June 16-20, 2019}, pages 9268--9277. Computer Vision Foundation / {IEEE}.

\bibitem[{Dettmers and Zettlemoyer(2023)}]{DBLP:conf/icml/DettmersZ23}
Tim Dettmers and Luke Zettlemoyer. 2023.
\newblock \href {https://proceedings.mlr.press/v202/dettmers23a.html} {The case for 4-bit precision: k-bit inference scaling laws}.
\newblock In \emph{International Conference on Machine Learning, {ICML} 2023, 23-29 July 2023, Honolulu, Hawaii, {USA}}, volume 202 of \emph{Proceedings of Machine Learning Research}, pages 7750--7774. {PMLR}.

\bibitem[{Devlin et~al.(2019)Devlin, Chang, Lee, and Toutanova}]{DBLP:conf/naacl/DevlinCLT19}
Jacob Devlin, Ming{-}Wei Chang, Kenton Lee, and Kristina Toutanova. 2019.
\newblock \href {https://doi.org/10.18653/v1/n19-1423} {{BERT:} pre-training of deep bidirectional transformers for language understanding}.
\newblock In \emph{Proceedings of the 2019 Conference of the North American Chapter of the Association for Computational Linguistics: Human Language Technologies, {NAACL-HLT} 2019, Minneapolis, MN, USA, June 2-7, 2019, Volume 1 (Long and Short Papers)}, pages 4171--4186. Association for Computational Linguistics.

\bibitem[{Dong et~al.(2019{\natexlab{a}})Dong, Yang, Wang, Wei, Liu, Wang, Gao, Zhou, and Hon}]{DBLP:conf/nips/00040WWLWGZH19}
Li~Dong, Nan Yang, Wenhui Wang, Furu Wei, Xiaodong Liu, Yu~Wang, Jianfeng Gao, Ming Zhou, and Hsiao{-}Wuen Hon. 2019{\natexlab{a}}.
\newblock \href {https://proceedings.neurips.cc/paper/2019/hash/c20bb2d9a50d5ac1f713f8b34d9aac5a-Abstract.html} {Unified language model pre-training for natural language understanding and generation}.
\newblock In \emph{Advances in Neural Information Processing Systems 32: Annual Conference on Neural Information Processing Systems 2019, NeurIPS 2019, December 8-14, 2019, Vancouver, BC, Canada}, pages 13042--13054.

\bibitem[{Dong et~al.(2019{\natexlab{b}})Dong, Gong, and Zhu}]{DBLP:journals/pami/DongGZ19}
Qi~Dong, Shaogang Gong, and Xiatian Zhu. 2019{\natexlab{b}}.
\newblock \href {https://doi.org/10.1109/TPAMI.2018.2832629} {Imbalanced deep learning by minority class incremental rectification}.
\newblock \emph{{IEEE} Trans. Pattern Anal. Mach. Intell.}, 41(6):1367--1381.

\bibitem[{Du et~al.(2022)Du, Huang, Dai, Tong, Lepikhin, Xu, Krikun, Zhou, Yu, Firat, Zoph, Fedus, Bosma, Zhou, Wang, Wang, Webster, Pellat, Robinson, Meier{-}Hellstern, Duke, Dixon, Zhang, Le, Wu, Chen, and Cui}]{DBLP:conf/icml/DuHDTLXKZYFZFBZ22}
Nan Du, Yanping Huang, Andrew~M. Dai, Simon Tong, Dmitry Lepikhin, Yuanzhong Xu, Maxim Krikun, Yanqi Zhou, Adams~Wei Yu, Orhan Firat, Barret Zoph, Liam Fedus, Maarten~P. Bosma, Zongwei Zhou, Tao Wang, Yu~Emma Wang, Kellie Webster, Marie Pellat, Kevin Robinson, Kathleen~S. Meier{-}Hellstern, Toju Duke, Lucas Dixon, Kun Zhang, Quoc~V. Le, Yonghui Wu, Zhifeng Chen, and Claire Cui. 2022.
\newblock \href {https://proceedings.mlr.press/v162/du22c.html} {Glam: Efficient scaling of language models with mixture-of-experts}.
\newblock In \emph{International Conference on Machine Learning, {ICML} 2022, 17-23 July 2022, Baltimore, Maryland, {USA}}, volume 162 of \emph{Proceedings of Machine Learning Research}, pages 5547--5569. {PMLR}.

\bibitem[{Francis and Kucera(1979)}]{francis1979brown}
W~Nelson Francis and Henry Kucera. 1979.
\newblock Brown corpus manual.
\newblock \emph{Letters to the Editor}, 5(2):7.

\bibitem[{Gao et~al.(2021{\natexlab{a}})Gao, Biderman, Black, Golding, Hoppe, Foster, Phang, He, Thite, Nabeshima, Presser, and Leahy}]{DBLP:journals/corr/abs-2101-00027}
Leo Gao, Stella Biderman, Sid Black, Laurence Golding, Travis Hoppe, Charles Foster, Jason Phang, Horace He, Anish Thite, Noa Nabeshima, Shawn Presser, and Connor Leahy. 2021{\natexlab{a}}.
\newblock \href {http://arxiv.org/abs/2101.00027} {The pile: An 800gb dataset of diverse text for language modeling}.
\newblock \emph{CoRR}, abs/2101.00027.

\bibitem[{Gao et~al.(2021{\natexlab{b}})Gao, Tow, Biderman, Black, DiPofi, Foster, Golding, Hsu, McDonell, Muennighoff, Phang, Reynolds, Tang, Thite, Wang, Wang, and Zou}]{eval-harness}
Leo Gao, Jonathan Tow, Stella Biderman, Sid Black, Anthony DiPofi, Charles Foster, Laurence Golding, Jeffrey Hsu, Kyle McDonell, Niklas Muennighoff, Jason Phang, Laria Reynolds, Eric Tang, Anish Thite, Ben Wang, Kevin Wang, and Andy Zou. 2021{\natexlab{b}}.
\newblock \href {https://doi.org/10.5281/zenodo.5371628} {A framework for few-shot language model evaluation}.

\bibitem[{Graves(2013)}]{DBLP:journals/corr/Graves13}
Alex Graves. 2013.
\newblock \href {http://arxiv.org/abs/1308.0850} {Generating sequences with recurrent neural networks}.
\newblock \emph{CoRR}, abs/1308.0850.

\bibitem[{Hendrycks et~al.(2021)Hendrycks, Burns, Basart, Zou, Mazeika, Song, and Steinhardt}]{DBLP:conf/iclr/HendrycksBBZMSS21}
Dan Hendrycks, Collin Burns, Steven Basart, Andy Zou, Mantas Mazeika, Dawn Song, and Jacob Steinhardt. 2021.
\newblock \href {https://openreview.net/forum?id=d7KBjmI3GmQ} {Measuring massive multitask language understanding}.
\newblock In \emph{9th International Conference on Learning Representations, {ICLR} 2021, Virtual Event, Austria, May 3-7, 2021}. OpenReview.net.

\bibitem[{Hoffmann et~al.(2022)Hoffmann, Borgeaud, Mensch, Buchatskaya, Cai, Rutherford, de~Las~Casas, Hendricks, Welbl, Clark, Hennigan, Noland, Millican, van~den Driessche, Damoc, Guy, Osindero, Simonyan, Elsen, Rae, Vinyals, and Sifre}]{DBLP:journals/corr/abs-2203-15556}
Jordan Hoffmann, Sebastian Borgeaud, Arthur Mensch, Elena Buchatskaya, Trevor Cai, Eliza Rutherford, Diego de~Las~Casas, Lisa~Anne Hendricks, Johannes Welbl, Aidan Clark, Tom Hennigan, Eric Noland, Katie Millican, George van~den Driessche, Bogdan Damoc, Aurelia Guy, Simon Osindero, Karen Simonyan, Erich Elsen, Jack~W. Rae, Oriol Vinyals, and Laurent Sifre. 2022.
\newblock \href {https://doi.org/10.48550/arXiv.2203.15556} {Training compute-optimal large language models}.
\newblock \emph{CoRR}, abs/2203.15556.

\bibitem[{Joshi et~al.(2017)Joshi, Choi, Weld, and Zettlemoyer}]{DBLP:conf/acl/JoshiCWZ17}
Mandar Joshi, Eunsol Choi, Daniel~S. Weld, and Luke Zettlemoyer. 2017.
\newblock \href {https://doi.org/10.18653/v1/P17-1147} {Triviaqa: {A} large scale distantly supervised challenge dataset for reading comprehension}.
\newblock In \emph{Proceedings of the 55th Annual Meeting of the Association for Computational Linguistics, {ACL} 2017, Vancouver, Canada, July 30 - August 4, Volume 1: Long Papers}, pages 1601--1611. Association for Computational Linguistics.

\bibitem[{Katz(1987)}]{DBLP:journals/tsp/Katz87}
Slava~M. Katz. 1987.
\newblock \href {https://doi.org/10.1109/TASSP.1987.1165125} {Estimation of probabilities from sparse data for the language model component of a speech recognizer}.
\newblock \emph{{IEEE} Trans. Acoust. Speech Signal Process.}, 35(3):400--401.

\bibitem[{Kneser and Ney(1995)}]{DBLP:conf/icassp/KneserN95}
Reinhard Kneser and Hermann Ney. 1995.
\newblock \href {https://doi.org/10.1109/ICASSP.1995.479394} {Improved backing-off for m-gram language modeling}.
\newblock In \emph{1995 International Conference on Acoustics, Speech, and Signal Processing, {ICASSP} '95, Detroit, Michigan, USA, May 08-12, 1995}, pages 181--184. {IEEE} Computer Society.

\bibitem[{Lee et~al.(2016)Lee, Park, and Kim}]{DBLP:conf/icip/LeePK16}
Hansang Lee, Minseok Park, and Junmo Kim. 2016.
\newblock \href {https://doi.org/10.1109/ICIP.2016.7533053} {Plankton classification on imbalanced large scale database via convolutional neural networks with transfer learning}.
\newblock In \emph{2016 {IEEE} International Conference on Image Processing, {ICIP} 2016, Phoenix, AZ, USA, September 25-28, 2016}, pages 3713--3717. {IEEE}.

\bibitem[{Lin et~al.(2017)Lin, Goyal, Girshick, He, and Doll{\'{a}}r}]{DBLP:conf/iccv/LinGGHD17}
Tsung{-}Yi Lin, Priya Goyal, Ross~B. Girshick, Kaiming He, and Piotr Doll{\'{a}}r. 2017.
\newblock \href {https://doi.org/10.1109/ICCV.2017.324} {Focal loss for dense object detection}.
\newblock In \emph{{IEEE} International Conference on Computer Vision, {ICCV} 2017, Venice, Italy, October 22-29, 2017}, pages 2999--3007. {IEEE} Computer Society.

\bibitem[{Liu et~al.(2019)Liu, Ott, Goyal, Du, Joshi, Chen, Levy, Lewis, Zettlemoyer, and Stoyanov}]{DBLP:journals/corr/abs-1907-11692}
Yinhan Liu, Myle Ott, Naman Goyal, Jingfei Du, Mandar Joshi, Danqi Chen, Omer Levy, Mike Lewis, Luke Zettlemoyer, and Veselin Stoyanov. 2019.
\newblock \href {http://arxiv.org/abs/1907.11692} {Roberta: {A} robustly optimized {BERT} pretraining approach}.
\newblock \emph{CoRR}, abs/1907.11692.

\bibitem[{Loshchilov and Hutter(2019)}]{DBLP:conf/iclr/LoshchilovH19}
Ilya Loshchilov and Frank Hutter. 2019.
\newblock \href {https://openreview.net/forum?id=Bkg6RiCqY7} {Decoupled weight decay regularization}.
\newblock In \emph{7th International Conference on Learning Representations, {ICLR} 2019, New Orleans, LA, USA, May 6-9, 2019}. OpenReview.net.

\bibitem[{Mihaylov et~al.(2018)Mihaylov, Clark, Khot, and Sabharwal}]{DBLP:conf/emnlp/MihaylovCKS18}
Todor Mihaylov, Peter Clark, Tushar Khot, and Ashish Sabharwal. 2018.
\newblock \href {https://doi.org/10.18653/v1/d18-1260} {Can a suit of armor conduct electricity? {A} new dataset for open book question answering}.
\newblock In \emph{Proceedings of the 2018 Conference on Empirical Methods in Natural Language Processing, Brussels, Belgium, October 31 - November 4, 2018}, pages 2381--2391. Association for Computational Linguistics.

\bibitem[{Mikolov et~al.(2010)Mikolov, Karafi{\'{a}}t, Burget, Cernock{\'{y}}, and Khudanpur}]{DBLP:conf/interspeech/MikolovKBCK10}
Tom{\'{a}}s Mikolov, Martin Karafi{\'{a}}t, Luk{\'{a}}s Burget, Jan Cernock{\'{y}}, and Sanjeev Khudanpur. 2010.
\newblock \href {https://doi.org/10.21437/Interspeech.2010-343} {Recurrent neural network based language model}.
\newblock In \emph{{INTERSPEECH} 2010, 11th Annual Conference of the International Speech Communication Association, Makuhari, Chiba, Japan, September 26-30, 2010}, pages 1045--1048. {ISCA}.

\bibitem[{Mostafazadeh et~al.(2016)Mostafazadeh, Chambers, He, Parikh, Batra, Vanderwende, Kohli, and Allen}]{DBLP:journals/corr/MostafazadehCHP16}
Nasrin Mostafazadeh, Nathanael Chambers, Xiaodong He, Devi Parikh, Dhruv Batra, Lucy Vanderwende, Pushmeet Kohli, and James~F. Allen. 2016.
\newblock \href {http://arxiv.org/abs/1604.01696} {A corpus and evaluation framework for deeper understanding of commonsense stories}.
\newblock \emph{CoRR}, abs/1604.01696.

\bibitem[{Paperno et~al.(2016)Paperno, Kruszewski, Lazaridou, Pham, Bernardi, Pezzelle, Baroni, Boleda, and Fern{\'{a}}ndez}]{DBLP:conf/acl/PapernoKLPBPBBF16}
Denis Paperno, Germ{\'{a}}n Kruszewski, Angeliki Lazaridou, Quan~Ngoc Pham, Raffaella Bernardi, Sandro Pezzelle, Marco Baroni, Gemma Boleda, and Raquel Fern{\'{a}}ndez. 2016.
\newblock \href {https://doi.org/10.18653/v1/p16-1144} {The {LAMBADA} dataset: Word prediction requiring a broad discourse context}.
\newblock In \emph{Proceedings of the 54th Annual Meeting of the Association for Computational Linguistics, {ACL} 2016, August 7-12, 2016, Berlin, Germany, Volume 1: Long Papers}. The Association for Computer Linguistics.

\bibitem[{Piantadosi(2014)}]{piantadosi2014zipf}
Steven~T Piantadosi. 2014.
\newblock Zipf’s word frequency law in natural language: A critical review and future directions.
\newblock \emph{Psychonomic bulletin \& review}, 21:1112--1130.

\bibitem[{Pouyanfar et~al.(2018)Pouyanfar, Tao, Mohan, Tian, Kaseb, Gauen, Dailey, Aghajanzadeh, Lu, Chen, and Shyu}]{DBLP:conf/mipr/PouyanfarTMTKGD18}
Samira Pouyanfar, Yudong Tao, Anup Mohan, Haiman Tian, Ahmed~S. Kaseb, Kent Gauen, Ryan Dailey, Sarah Aghajanzadeh, Yung{-}Hsiang Lu, Shu{-}Ching Chen, and Mei{-}Ling Shyu. 2018.
\newblock \href {https://doi.org/10.1109/MIPR.2018.00027} {Dynamic sampling in convolutional neural networks for imbalanced data classification}.
\newblock In \emph{{IEEE} 1st Conference on Multimedia Information Processing and Retrieval, {MIPR} 2018, Miami, FL, USA, April 10-12, 2018}, pages 112--117. {IEEE}.

\bibitem[{Radford et~al.(2019)Radford, Wu, Child, Luan, Amodei, Sutskever et~al.}]{radford2019language}
Alec Radford, Jeffrey Wu, Rewon Child, David Luan, Dario Amodei, Ilya Sutskever, et~al. 2019.
\newblock Language models are unsupervised multitask learners.
\newblock \emph{OpenAI blog}, 1(8):9.

\bibitem[{Rae et~al.(2021)Rae, Borgeaud, Cai, Millican, Hoffmann, Song, Aslanides, Henderson, Ring, Young, Rutherford, Hennigan, Menick, Cassirer, Powell, van~den Driessche, Hendricks, Rauh, Huang, Glaese, Welbl, Dathathri, Huang, Uesato, Mellor, Higgins, Creswell, McAleese, Wu, Elsen, Jayakumar, Buchatskaya, Budden, Sutherland, Simonyan, Paganini, Sifre, Martens, Li, Kuncoro, Nematzadeh, Gribovskaya, Donato, Lazaridou, Mensch, Lespiau, Tsimpoukelli, Grigorev, Fritz, Sottiaux, Pajarskas, Pohlen, Gong, Toyama, de~Masson~d'Autume, Li, Terzi, Mikulik, Babuschkin, Clark, de~Las~Casas, Guy, Jones, Bradbury, Johnson, Hechtman, Weidinger, Gabriel, Isaac, Lockhart, Osindero, Rimell, Dyer, Vinyals, Ayoub, Stanway, Bennett, Hassabis, Kavukcuoglu, and Irving}]{DBLP:journals/corr/abs-2112-11446}
Jack~W. Rae, Sebastian Borgeaud, Trevor Cai, Katie Millican, Jordan Hoffmann, H.~Francis Song, John Aslanides, Sarah Henderson, Roman Ring, Susannah Young, Eliza Rutherford, Tom Hennigan, Jacob Menick, Albin Cassirer, Richard Powell, George van~den Driessche, Lisa~Anne Hendricks, Maribeth Rauh, Po{-}Sen Huang, Amelia Glaese, Johannes Welbl, Sumanth Dathathri, Saffron Huang, Jonathan Uesato, John Mellor, Irina Higgins, Antonia Creswell, Nat McAleese, Amy Wu, Erich Elsen, Siddhant~M. Jayakumar, Elena Buchatskaya, David Budden, Esme Sutherland, Karen Simonyan, Michela Paganini, Laurent Sifre, Lena Martens, Xiang~Lorraine Li, Adhiguna Kuncoro, Aida Nematzadeh, Elena Gribovskaya, Domenic Donato, Angeliki Lazaridou, Arthur Mensch, Jean{-}Baptiste Lespiau, Maria Tsimpoukelli, Nikolai Grigorev, Doug Fritz, Thibault Sottiaux, Mantas Pajarskas, Toby Pohlen, Zhitao Gong, Daniel Toyama, Cyprien de~Masson~d'Autume, Yujia Li, Tayfun Terzi, Vladimir Mikulik, Igor Babuschkin, Aidan Clark, Diego de~Las~Casas, Aurelia Guy,
  Chris Jones, James Bradbury, Matthew~J. Johnson, Blake~A. Hechtman, Laura Weidinger, Iason Gabriel, William Isaac, Edward Lockhart, Simon Osindero, Laura Rimell, Chris Dyer, Oriol Vinyals, Kareem Ayoub, Jeff Stanway, Lorrayne Bennett, Demis Hassabis, Koray Kavukcuoglu, and Geoffrey Irving. 2021.
\newblock \href {http://arxiv.org/abs/2112.11446} {Scaling language models: Methods, analysis {\&} insights from training gopher}.
\newblock \emph{CoRR}, abs/2112.11446.

\bibitem[{Raffel et~al.(2020)Raffel, Shazeer, Roberts, Lee, Narang, Matena, Zhou, Li, and Liu}]{DBLP:journals/jmlr/RaffelSRLNMZLL20}
Colin Raffel, Noam Shazeer, Adam Roberts, Katherine Lee, Sharan Narang, Michael Matena, Yanqi Zhou, Wei Li, and Peter~J. Liu. 2020.
\newblock \href {http://jmlr.org/papers/v21/20-074.html} {Exploring the limits of transfer learning with a unified text-to-text transformer}.
\newblock \emph{J. Mach. Learn. Res.}, 21:140:1--140:67.

\bibitem[{Sakaguchi et~al.(2020)Sakaguchi, Bras, Bhagavatula, and Choi}]{DBLP:conf/aaai/SakaguchiBBC20}
Keisuke Sakaguchi, Ronan~Le Bras, Chandra Bhagavatula, and Yejin Choi. 2020.
\newblock \href {https://doi.org/10.1609/aaai.v34i05.6399} {Winogrande: An adversarial winograd schema challenge at scale}.
\newblock In \emph{The Thirty-Fourth {AAAI} Conference on Artificial Intelligence, {AAAI} 2020, The Thirty-Second Innovative Applications of Artificial Intelligence Conference, {IAAI} 2020, The Tenth {AAAI} Symposium on Educational Advances in Artificial Intelligence, {EAAI} 2020, New York, NY, USA, February 7-12, 2020}, pages 8732--8740. {AAAI} Press.

\bibitem[{Sap et~al.(2019)Sap, Rashkin, Chen, Bras, and Choi}]{DBLP:journals/corr/abs-1904-09728}
Maarten Sap, Hannah Rashkin, Derek Chen, Ronan~Le Bras, and Yejin Choi. 2019.
\newblock \href {http://arxiv.org/abs/1904.09728} {Socialiqa: Commonsense reasoning about social interactions}.
\newblock \emph{CoRR}, abs/1904.09728.

\bibitem[{Shen et~al.(2016)Shen, Lin, and Huang}]{shen2016relay}
Li~Shen, Zhouchen Lin, and Qingming Huang. 2016.
\newblock Relay backpropagation for effective learning of deep convolutional neural networks.
\newblock In \emph{Computer Vision--ECCV 2016: 14th European Conference, Amsterdam, The Netherlands, October 11--14, 2016, Proceedings, Part VII 14}, pages 467--482. Springer.

\bibitem[{Touvron et~al.(2023)Touvron, Lavril, Izacard, Martinet, Lachaux, Lacroix, Rozi{\`{e}}re, Goyal, Hambro, Azhar, Rodriguez, Joulin, Grave, and Lample}]{DBLP:journals/corr/abs-2302-13971}
Hugo Touvron, Thibaut Lavril, Gautier Izacard, Xavier Martinet, Marie{-}Anne Lachaux, Timoth{\'{e}}e Lacroix, Baptiste Rozi{\`{e}}re, Naman Goyal, Eric Hambro, Faisal Azhar, Aur{\'{e}}lien Rodriguez, Armand Joulin, Edouard Grave, and Guillaume Lample. 2023.
\newblock \href {https://doi.org/10.48550/arXiv.2302.13971} {Llama: Open and efficient foundation language models}.
\newblock \emph{CoRR}, abs/2302.13971.

\bibitem[{Tsoumakas and Katakis(2007)}]{tsoumakas2007multi}
Grigorios Tsoumakas and Ioannis Katakis. 2007.
\newblock Multi-label classification: An overview.
\newblock \emph{International Journal of Data Warehousing and Mining (IJDWM)}, 3(3):1--13.

\bibitem[{Vaswani et~al.(2017)Vaswani, Shazeer, Parmar, Uszkoreit, Jones, Gomez, Kaiser, and Polosukhin}]{DBLP:conf/nips/VaswaniSPUJGKP17}
Ashish Vaswani, Noam Shazeer, Niki Parmar, Jakob Uszkoreit, Llion Jones, Aidan~N. Gomez, Lukasz Kaiser, and Illia Polosukhin. 2017.
\newblock \href {https://proceedings.neurips.cc/paper/2017/hash/3f5ee243547dee91fbd053c1c4a845aa-Abstract.html} {Attention is all you need}.
\newblock In \emph{Advances in Neural Information Processing Systems 30: Annual Conference on Neural Information Processing Systems 2017, December 4-9, 2017, Long Beach, CA, {USA}}, pages 5998--6008.

\bibitem[{Xie et~al.(2023)Xie, Pham, Dong, Du, Liu, Lu, Liang, Le, Ma, and Yu}]{DBLP:journals/corr/abs-2305-10429}
Sang~Michael Xie, Hieu Pham, Xuanyi Dong, Nan Du, Hanxiao Liu, Yifeng Lu, Percy Liang, Quoc~V. Le, Tengyu Ma, and Adams~Wei Yu. 2023.
\newblock \href {https://doi.org/10.48550/arXiv.2305.10429} {Doremi: Optimizing data mixtures speeds up language model pretraining}.
\newblock \emph{CoRR}, abs/2305.10429.

\bibitem[{Yang and Xu(2020)}]{yang2020rethinking}
Yuzhe Yang and Zhi Xu. 2020.
\newblock Rethinking the value of labels for improving class-imbalanced learning.
\newblock \emph{Advances in neural information processing systems}, 33:19290--19301.

\bibitem[{Zellers et~al.(2019)Zellers, Holtzman, Bisk, Farhadi, and Choi}]{DBLP:conf/acl/ZellersHBFC19}
Rowan Zellers, Ari Holtzman, Yonatan Bisk, Ali Farhadi, and Yejin Choi. 2019.
\newblock \href {https://doi.org/10.18653/v1/p19-1472} {Hellaswag: Can a machine really finish your sentence?}
\newblock In \emph{Proceedings of the 57th Conference of the Association for Computational Linguistics, {ACL} 2019, Florence, Italy, July 28- August 2, 2019, Volume 1: Long Papers}, pages 4791--4800. Association for Computational Linguistics.

\bibitem[{Zhang et~al.(2022)Zhang, Roller, Goyal, Artetxe, Chen, Chen, Dewan, Diab, Li, Lin et~al.}]{zhang2022opt}
Susan Zhang, Stephen Roller, Naman Goyal, Mikel Artetxe, Moya Chen, Shuohui Chen, Christopher Dewan, Mona Diab, Xian Li, Xi~Victoria Lin, et~al. 2022.
\newblock Opt: Open pre-trained transformer language models.
\newblock \emph{arXiv preprint arXiv:2205.01068}.

\end{thebibliography}

\end{document}